\documentclass[preprint,12pt]{elsarticle}
\usepackage{graphicx}
\usepackage{sectsty}
\usepackage{amsmath}

\usepackage{amssymb}

\usepackage{siunitx}
\usepackage{nicefrac}
\newcommand{\ytrue}{\ensuremath{y_{\text{true}}}}
\newcommand{\ypred}{\ensuremath{y_{\text{pred}}}}

\journal{Engineering Applications of Artificial Intelligence}

\begin{document}

\begin{frontmatter}



\title{Using Supervised Deep-Learning to Model Edge-FBG Shape Sensors}

\author[inst1]{Samaneh Manavi Roodsari}
\author[inst1]{Antal Huck-Horvath}
\author[inst1]{Sara Freund}
\affiliation[inst1]{organization={Center for medical Image Analysis and Navigation (CIAN), Department of Biomedical Engineering, University of Basel},
            addressline={Gewerbestrasse 14}, 
            city={Allschwil},
            postcode={4123}, 
            country={Switzerland}}
\author[inst2,inst3]{Azhar Zam}            
\affiliation[inst2]{organization={Division of Engineering, New York University Abu Dhabi},
            city={Abu Dhabi},
            postcode={129188}, 
            country={UAE}}
\affiliation[inst3]{organization={Tandon School of Engineering, New York University},
            city={Brooklyn, NY},
            postcode={11201}, 
            country={USA}}
\author[inst4]{Georg Rauter}  
\affiliation[inst4]{organization={Bio-Inspired RObots for MEDicine-Laboratory (BIROMED-Lab), Department of Biomedical Engineering, University of Basel},
            country={Switzerland}}
\author[inst5]{Wolfgang Schade}  
\affiliation[inst5]{organization={Department of Fiber Optical Sensor Systems, Fraunhofer Institute for Telecommunications, Heinrich Hertz Institute, HHI},
            addressline={Am Stollen 19H}, 
            city={Goslar},
            postcode={38640}, 
            country={Germany}}
\author[inst1]{Philippe C. Cattin}            

\begin{abstract}
Continuum robots in robot-assisted minimally invasive surgeries provide adequate access to target anatomies that are not directly reachable through small incisions. Achieving precise and reliable motion control of such snake-like manipulators necessitates an accurate navigation system that requires no line-of-sight and is immune to electromagnetic noises. 
Fiber Bragg Grating (FBG) shape sensors, particularly edge-FBGs, are promising tools for this task. However, in edge-FBG sensors, the intensity ratio between Bragg wavelengths carries the strain information that can be affected by undesired bending-related phenomena, making standard characterization techniques less suitable for these sensors. We showed in our previous work that a deep learning model has the potential to extract the strain information from the full edge-FBG spectrum and accurately predict the sensor's shape.
In this paper, we conduct a more thorough investigation to find a suitable architectural design with lower prediction errors. We use the Hyperband algorithm to search for optimal hyperparameters in two steps. First, we limit the search space to layer settings, where the best-performing configuration gets selected. Then, we modify the search space for tuning the training and loss calculation hyperparameters. We also analyze various data transformations on the input and output variables, as data rescaling can directly influence the model's performance. Moreover, we performed discriminative training using the Siamese network architecture that employs two CNNs with identical parameters to learn similarity metrics between the spectra of similar target values. The best-performing network architecture among all evaluated configurations can predict the sensor's shape with a median tip error of \SI{3.11}{\mm}.

\end{abstract}




\end{frontmatter}


\section{Introduction}
\label{sec:Introduction}

Minimally invasive surgical procedures (MIS) are delicate operations performed through small incisions or natural orifices on anatomical structures of the human body. Such interventions are beneficial compared to conventional open surgeries, as they reduce patient trauma, shorten recovery time \cite{burgner2015continuum}, and ensure overall cost-effectiveness \cite{vitiello2012emerging}. In addition, surgical robots help to realize the full potential of MIS procedures by enhancing dexterity and manipulability, as well as improving stability and motion accuracy \cite{vitiello2012emerging}. Continuum robots play an important role in complex robot-assisted MIS procedures, where no adequate and direct access through small incisions to target anatomies is available \cite{burgner2015continuum, dogangil2010review, van2014design, wang2012robotics, yeung2016application, payne2014hand, patel2014flexible}. However, achieving precise and reliable motion control of continuum robots requires accurate and real-time shape sensing. Accurately modeling the shape of these instruments remains challenging due to their inherent snake-like design and inevitable collisions with the surrounding tissues during the surgery \cite{webster2010design, xu2010analytic}. Therefore, a precise and accurate tracking system is needed to enable closed-loop control for such flexible manipulators.

The most common and commercially available medical tracking systems include optical tracking systems \cite{OpticalTracking,OpticalTracking_2}, electromagnetic sensors \cite{OpticalTracking}, intraoperative imaging technologies \cite{wagner20164d}, angular sensors \cite{glossop2012localization, peters2008image}, and FBG-based sensors \cite{glossop2012localization}.
Optical trackers are state-of-the-art technology for tracking medical tools and patients inside the operating room (OR). The optical trackers consist of cameras that detect navigation markers attached to the object of interest. They can navigate up to 25 tools with sub-millimeter accuracy over a large measurement volume (\textit{e.g.,} Polaris NDI $\sim$\,\SI{0.12}{\mm} RMS (root-mean-square) over a volume of $\sim$\,\SI{2}{\cubic\m} \cite{OpticalTracking}). Wireless tracking, reliable measurement, and stable performance are the other key advantages of this technology. However, they require a line-of-sight and are best suited to use with large, rigid tools. These substantial limitations make optical trackers unsuitable for navigating flexible endoscopes inside the patient's body. 

Electromagnetic (EM) tracking systems consist of two key components, field generators and wired sensor coils. The field generator emits a defined low-intensity EM field which establishes the measurement volume (\textit{e.g.,} AURORA NDI, Planar 20-20 FG \SI{75}{\cubic\cm} \cite{OpticalTracking}). Once the sensors enter the measurement volume, a small current is induced inside them, which is then used to determine the position and orientation of the sensor relative to the patient. This technology allows intracorporeal tracking, as it does not require a line-of-sight and can be embedded or placed at the tip of flexible tools. However, EM tracking systems are less accurate than optical trackers and have a smaller working volume. Adding multiple sensors along the endoscope is often impossible, as the sensors must be wired. Furthermore, they are sensitive to environmental EM interferences (\textit{e.g.,} the EM field of the robot) and to the presence of conductive or ferromagnetic metals. 

Intraoperative imaging modalities, including fluoroscopy, cone-beam CT, and ultrasound, can be an alternative to EM sensors for intracorporeal tracking. Some imaging modalities like biplane fluoroscopy achieve even higher accuracy as compared to EM sensors (mean shape error of \SI{0.54}{\mm} \cite{wagner20164d}) but are challenging to perform in crowded OR settings. In addition, they have limitations such as high doses of radiation (\textit{e.g.,} X-ray-based imaging), high computational cost (\textit{e.g.,} cone-beam CT), and low resolution (\textit{e.g.,} ultrasound).

In FBG-based shape sensors, the main components are a sensing probe (coated optical fibers) and an interrogation system for measuring the sensor's signal. FBG sensors can track themselves in three dimensions, thus providing real-time feedback on the shape and tip location when inserted into flexible instruments. These sensors are easily integrable into medical devices for tracking, as one single fiber with a typical diameter of \SI{250}{\micro\metre} can carry an array of FBGs to extract strain information along the length of the fiber. Moreover, FBG sensors are immune to EM interferences and are applicable for navigating robotic tools. The coating layer of the optical fiber can be a bio-compatible material, which makes it suitable for tracking catheters as well \cite{ryu2014fbg}.

Although different configurations for FBG-based shape sensors have been studied in recent years \cite{ryu2014fbg, manavi2018temperature, manavi2021using, roesthuis2013using, jackle2019fiber, marowsky2014planar, yi2012spatial}, the only fiber shape sensors that have been commercialized work based on multicore fibers (\textit{e.g.,} \cite{Sensuron2022, ShapeSensing}). Multicore fiber shape sensors are able to track themselves with a millimeter range accuracy (\textit{e.g.,} an average error of \SI{1.13}{\mm} for a \SI{38}{\cm} long sensor \cite{jackle2019fiber}). However, the cost of such systems is quite high \cite{waltermann2015femtosecond}, as an optical frequency domain reflectometer (OFDR) is needed for interrogating the FBGs \cite{soller2005high}, plus a fan-out device for reading the signal from each core \cite{thomson2007ultrafast}. 
Recently, a new configuration for FBG-based shape sensing, called edge-FBG, has been proposed by Waltermann \textit{et al.} \cite{waltermann2018multiple}, in which the FBGs are inscribed on the edge of the core in a single-mode optical fiber. Unlike standard FBG-based shape sensors, edge-FBG sensors are intensity-based, and the strain information is carried by the intensity ratio between the Bragg wavelengths. Such sensors can be interrogated using a broadband light source and a low-cost spectrometer, making the edge-FBG sensor a suitable choice for many applications where the price should be relatively low.

As explained in \cite{waltermann2018multiple}, the mode field's center in a single-mode fiber is sensitive to shape deformations and moves towards the opposite direction of bending. Depending on the radial and the angular distance between the dislocated mode field and the edge-FBGs, the intensity ratio between the edge-FBGs at each sensing plane changes. In this method, the curvature and the bending direction can be calculated from the estimated mode field centroid, and a shape reconstruction accuracy of $\sim$\SI{5}{\cm} for a \SI{25}{\cm} long sensor can be achieved~\cite{roodsari2022secretarxive}. However, in such eccentric FBGs, the intensity values at the Bragg wavelengths also depend on the spectrum profile of the incident light. Macro bending in optical fibers may affect the spectrum profile by causing wavelength-dependent attenuation \cite{faustini1997bend}. Moreover, in curved areas of an optical fiber, the refractive index profile is asymmetric (known as bending-induced birefringence), and therefore, changes in the light's polarization state are wavelength-dependent \cite{drexler2011optical}. Consequently, polarization-sensitive elements inside the FBG interrogation system may modify the spectrum profile by attenuating each wavelength element differently and cause errors in intensity measurements.

The authors showed in \cite{manavi2021using} that considering the complicated impact of bending-induced phenomena on the signal of edge-FBG sensors, it is feasible to model such sensors using deep learning techniques that predict the sensor's shape based on the full spectrum and not only the intensity at Bragg wavelengths. Meaning that it is already considering the effect of bending-related phenomena on the spectrum profile.

In this paper, we investigate the usage of deep learning algorithms for modeling edge-FBG sensors in more detail. First, we focus on identifying a good set of tuning parameters, known as hyperparameters, for our deep learning algorithm to extract relevant features from the edge-FBG sensor. We perform this hyperparameter tuning when the model's input (the sensor's spectra) and output data (the sensor's spatial shape) are preprocessed using different rescaling methods. Ultimately, we employ the most suitable data rescaling approach and the optimized feature-extracting network to perform discriminative training using the Siamese network~\cite{bromley1993signature}.

\section{Methodology}
\label{sec:Methodology}

The importance of choosing a good set of hyperparameters for a deep learning algorithm is well-known. The Hyperband, as one of the most common hyperparameter optimizers, considers several possible resource allocations (\textit{e.g.,} the total number of epochs used during evaluation) and invokes Successive Halving \cite{jamieson2016non} on randomly sampled hyperparameter configurations \cite{li2017hyperband}. Compared to black-box approaches like Bayesian optimization, the Hyperband is 5$\times$ to 30$\times$ faster and evaluates an order-of-magnitude more configurations \cite{li2017hyperband}. 
In addition to the hyperparameter configuration, rescaling input and output variables before being presented to deep neural networks greatly affects the model's performance \cite{bishop1995neural}. When input variables have large values, the model learns large weights, which may cause numerical instability, poor performance during training, and generalization error in the network. Large-scaled target values result in significant update rates in weight values, making the learning process unstable. In common practices, preprocessing transformations are applied to input variables prior to training the networks, and postprocessing steps are introduced to the model's predictions for calculating the desired target values \cite{bishop1995neural}. Therefore, we investigate different rescaling methods on the input and output variables to identify the most suited data preprocessing steps.

\subsection{Training Setup}\label{TrainingSetup}
\begin{figure}[!t]
    \centering
    \includegraphics[width=0.8\textwidth]{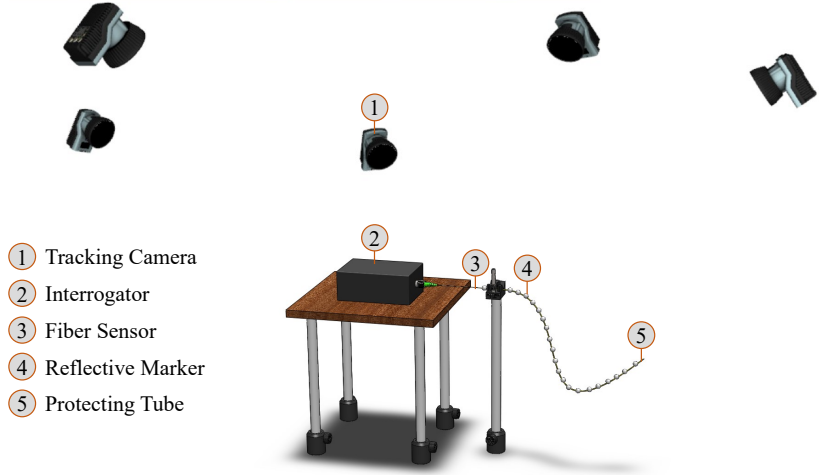}
    \caption{The data acquisition experimental setup of the edge-FBG sensor. Five tracking cameras (Oqus 7+) are used to monitor the sensor's shape. The sensor is inserted into a Hytrel furcation tubing to protect the fiber during shape manipulation.}
    \label{fig:setup}
\end{figure}

The dataset used in this work is from \cite{manavi2021using} with almost 53000 samples. Each sample consists of three consecutively measured edge-FBG spectra, the intensity values of 125 wavelength elements from \SI{812}{\nm} to \SI{871}{\nm}, as the input and the spatial coordinates of 21 discrete points along the fiber's length as corresponding target values. These discrete points are the positions of reflective markers that we attach to the sensor and monitor using a motion capture system. Figure~\ref{fig:setup} shows a schematic of the experimental setup. The FBG spectrum contains the reflected signal of 15 edge-FBGs from five sensing planes. Each sensing plane has three co-located FBGs at the left, top, and right edge of a single-mode fiber's core (SM800p from FIBERCORE company, UK).

\begin{figure}[!b]
    \centering
    \includegraphics[width=0.95\textwidth]{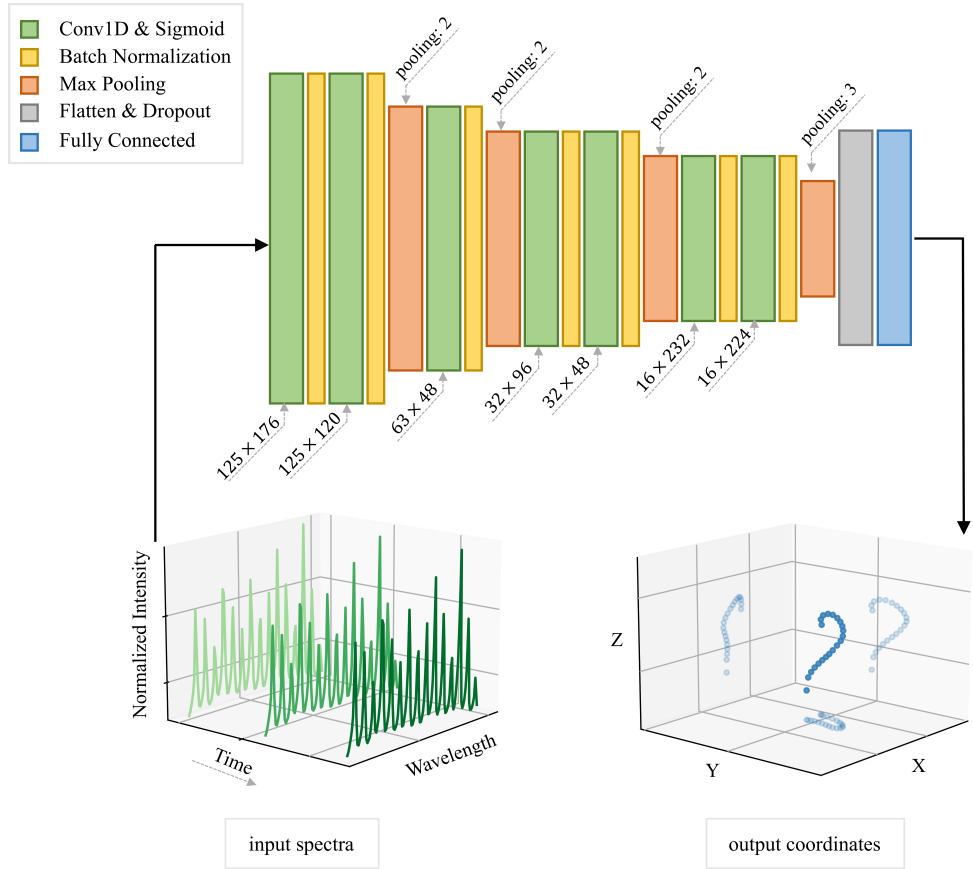}
    \caption{The architecture of the best-performing configuration after the first step of hyperparameter search. For this network, the input data is three consecutive edge-FBG spectra, normalized using one-dimensional z-scaling (see Section~\ref{InputDataPreprocessing} for more details), and the output data is the absolute coordinates of 21 discrete points along the sensor's length. The channel sizes of the seven 1D convolution layers (conv1D) are 176, 120, 48, 96, 48, 232, and 224, respectively. The pooling sizes of the four max pooling layers are 2, 2, 2, and 3, respectively.}
    \label{fig:Architecture}
\end{figure}

We use the Hyperband algorithm, built in the Keras tuner \cite{omalley2019kerastuner}, to perform hyperparameter optimization in two steps. First, a bigger search space is defined to optimize the settings of 1D convolutional layers (conv1D) and pooling layers (search criteria can be found in Table~\ref{tab:Searchingcriteria1} in the appendix). For this hyperparameter tuning step, the number of conv1D layers is set to seven. Based on our observations when investigating various search criteria, the network with seven conv1D is deep enough for feature extraction. Then, we fix these layer settings and modify the search space to tune the loss function, the optimizer, and the dropout rate (more detail on the search criteria is available in Table~\ref{tab:Searchingcriteria2} in the appendix). The model's input and output data are preprocessed using various data rescaling methods in this hyperparameter tuning step. The objective of the Hyperband is set to be the root-mean-square error on the validation set so that the scale of the scores, assigned to suggested configurations, is not affected when different loss functions get selected. As the Hyperband is based on a random search, we repeat each hyperparameter search three times for statistical robustness before selecting the final settings. To evaluate the predictive performance of the proposed model in an unbalanced way, the dataset is split into mutually disjoint Train-Validation-Test subsets: 80$\%$ for training, 10$\%$ for validating, and 10$\%$ for testing.

We investigated various configurations in the first hyperparameter tuning step. Figure~\ref{fig:Architecture} shows the best-performing architecture among the suggested configurations, in which each conv1D layer is followed by a Sigmoid activation function and batch normalization. The kernel size for the conv1D layers is $10$. Four max pooling layers are placed after the conv1D layers number two, three, five, and seven for down sampling the features. The final layer is a fully connected layer with a linear activation function to map the extracted features into desired target values. These hyperparameters are fixed for the remainder of this paper.

\subsection{Input Data Preprocessing}\label{InputDataPreprocessing}
\begin{figure}[ht]
    \centering
    \includegraphics[width=1\textwidth]{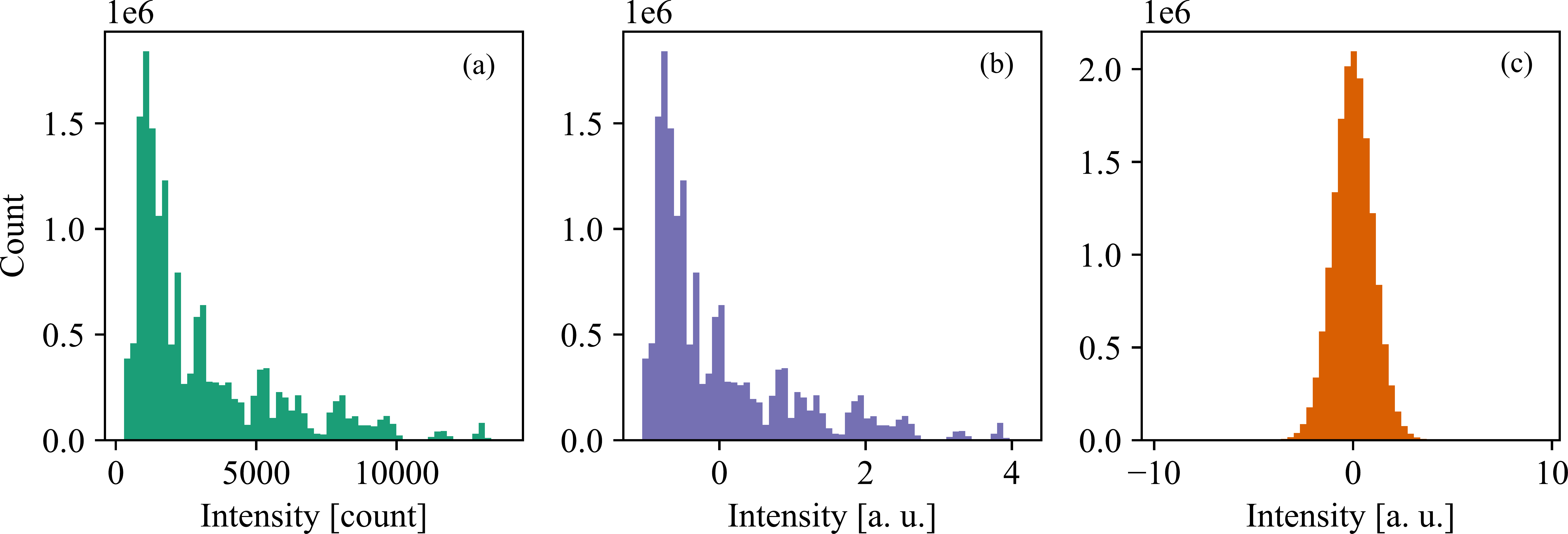}
    \caption{The histogram of the flat original input data (a), normalized input data using one-dimensional (b), and multi-dimensional z-scaling (c). The density distribution when using multi-dimensional z-scaling is approximately Gaussian, with skewness of $\sim 0.003$ and kurtosis of $\sim -1.2$. }
    \label{fig:Input Data Histogram}
\end{figure}

As mentioned earlier, the intensity ratios between Bragg wavelengths of co-located FBGs carry the strain information. Therefore, the input variables should not be normalized/standardized independently. We investigate two preprocessing transformations on the input variables, one-dimensional and multi-dimensional z-scaling~\cite{kessy2018optimal}. In the first method, we apply the standard scaling technique, considering the input data as a one-dimensional vector. The data distribution after rescaling has a zero mean value, and its standard deviation is one. Figure~\ref{fig:Input Data Histogram}(a) and (b) show the data distribution before and after rescaling. In the second method, we apply multi-dimensional standard scaling~\cite{kessy2018optimal} by subtracting each wavelength element from its mean value and dividing them over the square root of the covariance matrix

\begin{equation}
Z=U D^{-1/2} U^t (X - \mu),
\end{equation}
where $U$ and $D$ are the Eigenvectors and the Eigenvalues of the covariance matrix, $X$ is the input data (sensor's spectra), and $\mu$ is the mean intensity value at each wavelength element over the training dataset. With this approach, we achieve an approximately Gaussian density distribution (see Figure~\ref{fig:Input Data Histogram}(c)).

\subsection{Output Data Preprocessing} \label{OutputDataPreprocessing}

\begin{figure}[!b]
    \centering
    \includegraphics[width=0.8\textwidth]{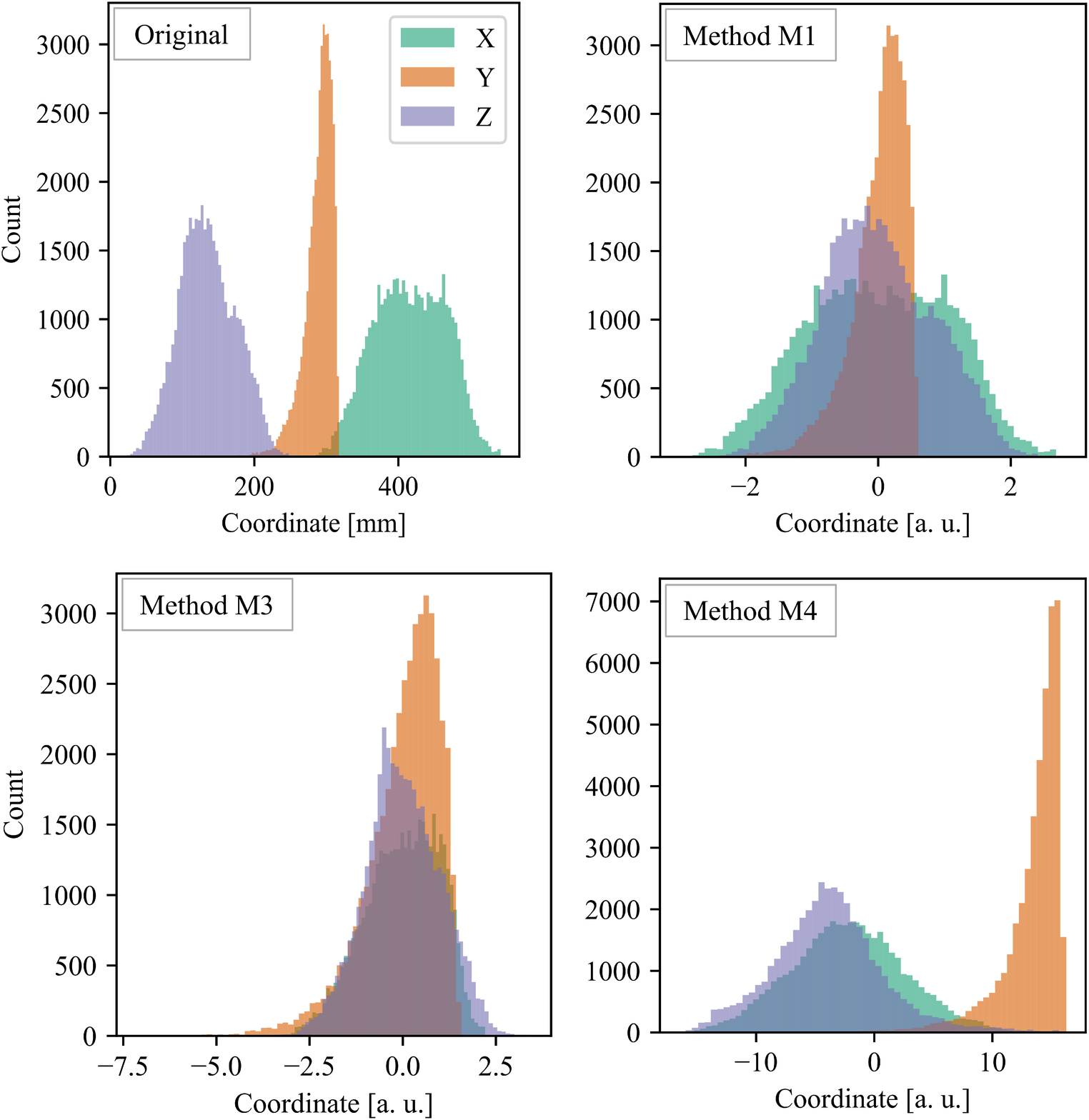}
    \caption{The histograms of the original $x$-, $y$-, and $z$-coordinates, their normalized values using M1 and M3, and the relative values using M4 for the point cloud number 10 are shown. The histogram of processed values using methods M2 is not depicted, as the distributions of $x$, $y$, and $z$ coordinates are similar to M1.}
    \label{fig:Out Data Histogram}
\end{figure}

In this section, we consider the measured coordinates of each marker as a point cloud and investigate the model's performance when various preprocessing transformations are applied to these point clouds (for more details, see the appendix).
In the first method (M1), we translate each point cloud to the origin of the global frame by subtracting the mean coordinate values from the original coordinates. We calculate the radial distance of the points from the cloud's center in all markers and compute its mean value. We then divide the translated coordinates by this calculated mean radial distance. In this method, the spatial coordinate elements for each marker have different scaling from the original data, but their density profile remains unchanged (see Figure~\ref{fig:Out Data Histogram}).
In the second approach (M2), each already translated point cloud is rescaled based on the mean of the radial distance of its own points, and therefore, the scaling factor is different for each point cloud. 
For the third method (M3), we apply a three-dimensional standard scaling to each point cloud. This way, the transformed data is uncorrelated, and the density profiles are different from the original ones. 
Lastly, as the fourth method (M4), we use the relative coordinates between the markers instead of absolute coordinate values. Figure~\ref{fig:Out Data Histogram} shows how the distribution of the coordinate elements changes when applying different preprocessing transformations to the target values.

\subsection{Siamese Network}
A further improvement of the network's performance was possible by guiding the feature-extracting part of the network in selecting relevant features for a given spectral sample \cite{wiatowski2017mathematical, mallat2012group}. The Siamese network \cite{bromley1993signature} is an architecture, designed for learning similarity metrics, which is well known from face recognition \cite{chopra2005learning,koch2015siamese} and handwritten forgery detection applications \cite{bromley1993signature,dey2017signet}. A Siamese network usually takes two inputs, compares them in the feature space, and provides a similarity measure between the two feature vectors. Siamese architectures consist of two identical subnetworks (the feature extractors) with shared weights that are trained using paired samples corresponding to similar (genuine) or dissimilar (imposter) outputs/pairs. During training, the feature extractor subnetworks are forced to provide vectors close to each other when the inputs belong to the same groups and are far away from each other if they are from different groups.

\begin{figure}[!t]
    \centering
    \includegraphics[width=0.95\textwidth]{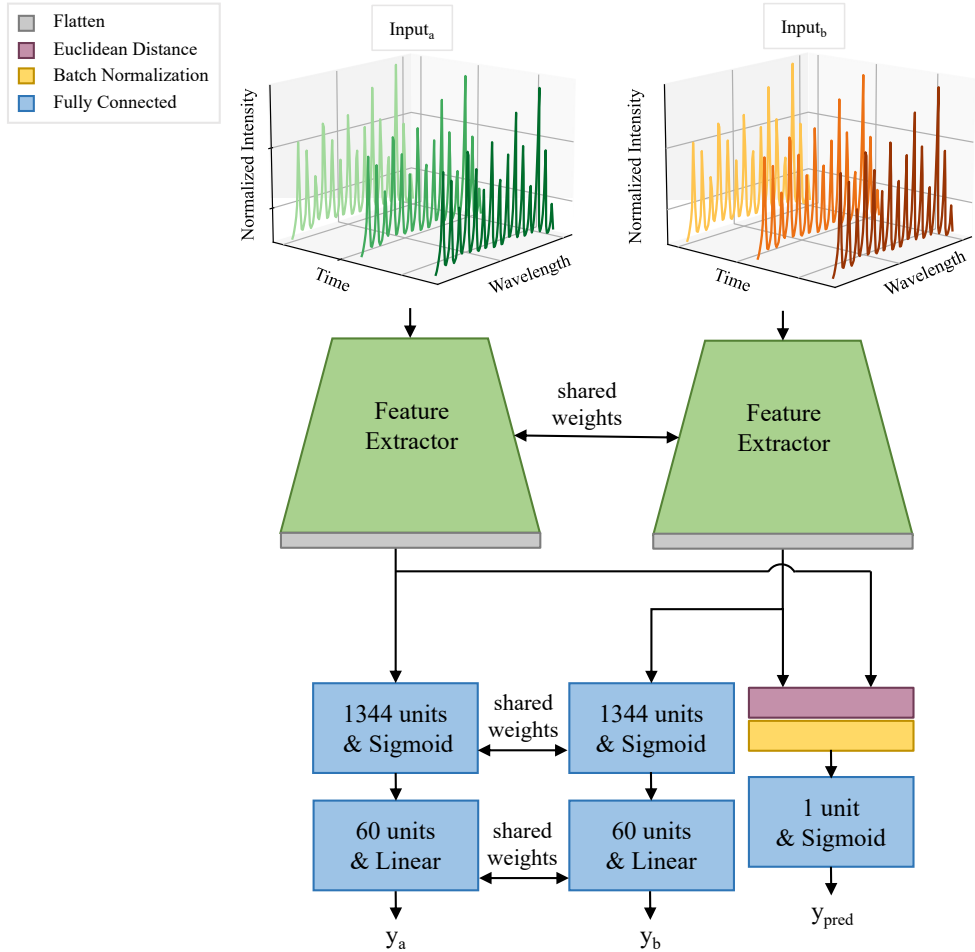}
    \caption{The architecture of the Siamese model. The feature extractor subnetworks have a similar architecture to what is shown in Figure~\ref{fig:Architecture}, but without the dropout and the fully connected layers.}
    \label{fig:Siamese}
\end{figure}

In our implementation, shown in Figure~\ref{fig:Siamese}, we use the same layer settings for feature extraction as in the architecture explained in Section~\ref{TrainingSetup}. We calculate the Euclidean distance between the two feature vectors, apply batch normalization, and pass it through a single-unit fully connected layer followed by a Sigmoid activation function. The output of the Sigmoid activation function gives a value close to one for distant feature vectors and a value close to zero for close vectors. In parallel, the feature vectors are also passed into two fully connected layers. The first one has $1344$ units followed by a Sigmoid activation function, and the second one is similar to the last layer of the previous architecture (compare Figure~\ref{fig:Architecture}), in which the preprocessed coordinates are calculated.
We group the samples using the RMSE, that is, the root-mean-square of the Euclidean distance between the corresponding shapes. First, we calculate the RMSE for all possible pairs in the training dataset. We then define the $1^{\text{\tiny st}}$ and the $25^{\text{\tiny th}}$ percentiles in the RMSE's histogram as thresholds for labeling the samples. These limits are selected after studying various combinations. If the calculated RMSE between two samples is less than the lower limit, we label the samples as zero (genuine pairs), and if it is within a $1\%$ range around the upper limit, we label them as one (imposter pairs).

The following loss function is used for this Siamese network:
\begin{equation}\label{eq:loss}
    \begin{split}
        \text{Loss} =\, \text{mean}\Big( & \,\alpha\left((1 - \ytrue)\, \ypred^2  + \ytrue\, \text{max}(0, M - \ypred)^2\right) \\
        &+(1-\alpha)\big(L(y_A-y_a) + L(y_B-y_b)\big)\Big)
    \end{split}
\end{equation}
where,
\begin{equation}
    \begin{split}
         L(a)=
        \begin{cases}
             0.5\, \frac{a^2}{\delta} & |a|\leqslant\delta,\\
             0.5\, \delta + (|a|-\delta) & \text{otherwise}.
        \end{cases}
    \end{split}
\end{equation}

In definition \eqref{eq:loss}, $\alpha$ is a scalar coefficient to weight loss contributions of the three model's outputs, and $\delta$ is a hyperparameter defining the range for mean-absolute-error, and mean-square-error in $L$, a modified version of Huber loss function \cite{huber1992robust}. $M$ is the margin, $\ytrue$ is the true label of paired samples, $\ypred$ is the output of the right arm, $y_A$ and $y_B$ are the true relative coordinates, and $y_a$ and $y_b$ are the predicted relative coordinates in the left and the middle arms of the network (Figure~\ref{fig:Siamese}).
The partial loss, calculated from $\ypred$, is the Contrastive loss~\cite{hadsell2006dimensionality}. Depending on whether the inputs are a genuine/imposter pair, the first or the second part of the Contrastive loss is applied to the output of the network's right arm. The two $L$ loss functions are calculated from predicted relative coordinate values in the left and middle arms of the network (Figure~\ref{fig:Siamese}). 
For a genuine pair, the network pushes $\ypred$ towards zero, such that the first part of the Contrastive loss also gets closer to zero. If the inputs are an imposter pair, $\ypred$ is forced to be larger than the value of $M$, such that the feature vectors stay separated in the feature space.
Similar to the data preprocessing Sections \ref{InputDataPreprocessing} \& \ref{OutputDataPreprocessing}, the training hyperparameters for this network, including the optimizer's parameters, $M$, $\alpha$, and $\delta$, were tuned using the Hyperband algorithm (see Table~\ref{tab:Searchingcriteria3} in the appendix for the search criteria). We ran the Hyperparameter search multiple times and selected the best-performing architecture for final training. 

\section{Results}
\label{sec:Results}

Evaluating the performance of the configurations suggested by the Hyperband is done by calculating the shape evaluation metrics between the true and the predicted shapes. Shape evaluation metrics include the tip error (the Euclidean distance between the true and the predicted coordinate of the sensor's tip) and the RMSE. The best-performing architecture among the three hyperparameter search attempts for each normalization method is selected based on the median values of the shape evaluation metrics in the validation dataset.

\begin{figure}[!b]
    \centering
    \includegraphics[width=0.8\textwidth]{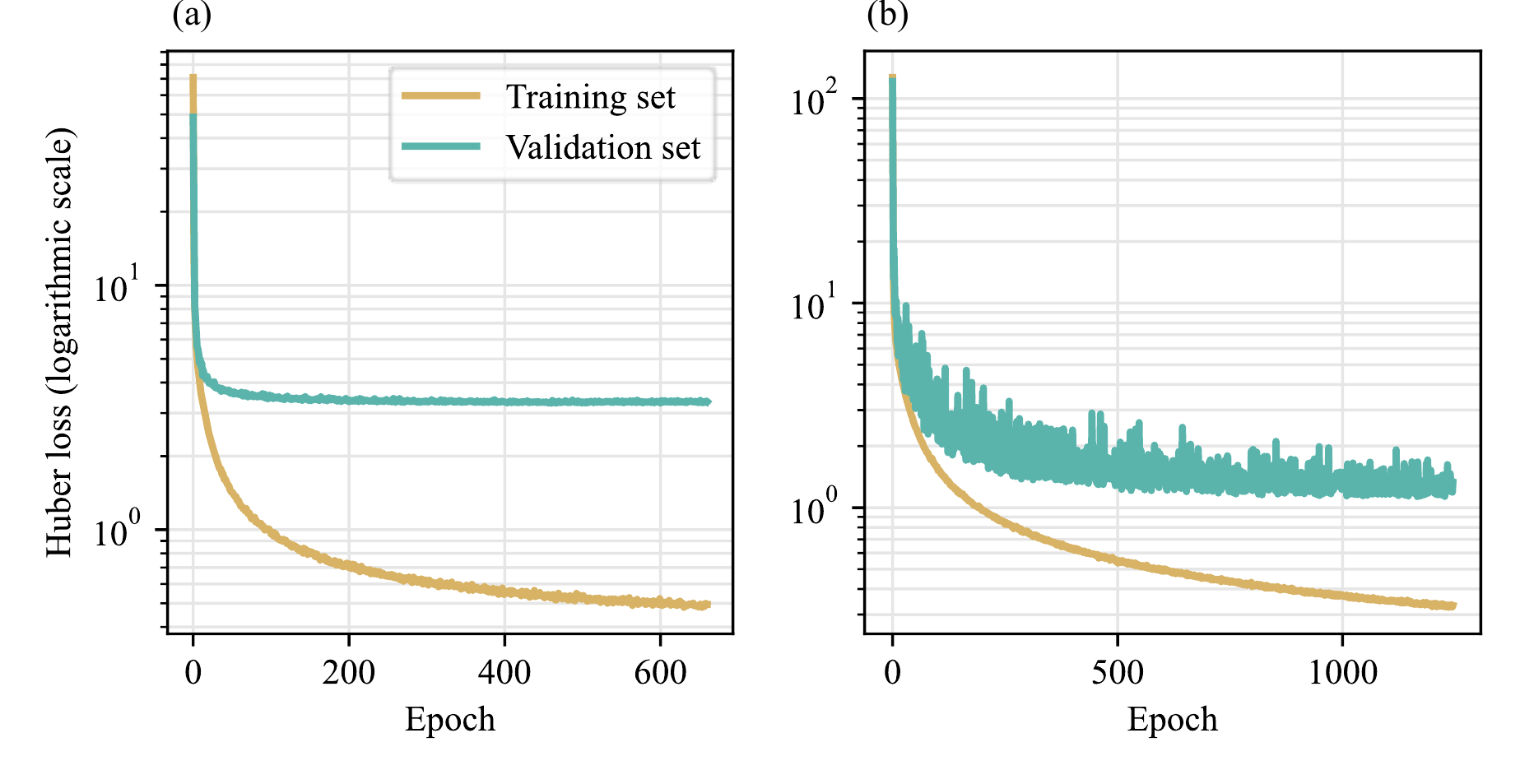}
    \caption{Training histories when the input data is preprocessed using (a) multi-dimensional and (b) one-dimensional normalization methods. The model was trained with early stopping conditions.}
    \label{fig:Training History}
\end{figure}

Table~\ref{tab:Input Data Error Table} shows the error values for the two input normalization methods. The one-dimensional normalization method, which preserves the distribution profile of the input data, results in a median tip error and a median RMSE of \SI{4.46}{\mm} and \SI{2.74}{\mm}, respectively. In the multi-dimensional normalization approach, the median values are \SI{13.38}{\mm} and \SI{8.11}{\mm}, respectively, which are significantly higher compared to the one-dimensional normalization method. This might be due to the validation loss reaching its plateau quicker when the input data distribution is approximately Gaussian (shown in Figure~\ref{fig:Training History}). The one-dimensional normalization method is therefore selected as the input data preprocessing step for the remainder of this paper. 

\begin{table}[!t]
    \fontsize{10}{12}\selectfont
    \caption{Shape evaluation errors on the test dataset when the input data are preprocessed using one-dimensional and multi-dimensional normalization methods. IQR: interquartile, one-dim.: one-dimensional normalization, multi-dim.: multi-dimensional normalization.}
    \begin{center}
    \begin{tabular}{lcccc}
    \hline
{\textbf{ }} & \multicolumn{2}{c}{\textbf{tip error\,{[}mm{]}}} & \multicolumn{2}{c}{\textbf{RMSE\,{[}mm{]}}} \\ 
\text{method} &\text{median} &\text{IQR} &\text{median} &\text{IQR} \\ \hline
\textbf{one-dim.}  &4.46   &4.30   &2.74  &2.39\\
\textbf{multi-dim.}  &13.38   &11.34   &8.11   &5.76\\
\hline
    \label{tab:Input Data Error Table}
    \end{tabular}
    \end{center}
\end{table}

The error statistics for shape evaluation metrics when using different preprocessing methods on the target data are shown in Table~\ref{tab:TargetData}.
Among the three preprocessing approaches applied to the absolute coordinate values (M1, M2, and M3), the first method (M1) shows the lowest error values with a median tip error of \SI{4.68}{\mm} and a median RMSE of \SI{2.83}{\mm}. However, the network performs better in predicting relative coordinates (M4) compared to absolute values by resulting in the median value of \SI{4.36}{\mm} for the tip error and \SI{2.78}{\mm} for the RMSE. Therefore, the relative coordinate values are selected as the best-performing target data preprocessing. 
The last row in Table~\ref{tab:TargetData} shows the shape evaluation parameters of the Siamese network when using the best-performing data preprocessing on both input and target data, one-dimensional normalization, and M4. As can be noted, there is a significant improvement in all error values, and the median tip error is reduced by almost \SI{1.25}{\mm} to \SI{3.11}{\mm} compared to M4 (more information on the significance test is provided in the appendix). The median value of RMSE is also reduced to \SI{1.98}{\mm} compared to the M4 method, which is \SI{2.78}{\mm}. 

\begin{table}[!t]
    \fontsize{10}{12}\selectfont
    \caption{Shape evaluation errors on the test dataset when the target data are processed using four different methods and when the network architecture is modified based on the Siamese design (indicated in bold). The model's output is first scaled back to absolute coordinates for each method, and then the error values are computed.}
    \label{tab:TargetData}
    \begin{center}
    \begin{tabular}{lcccc}
    \hline
{\textbf{ }} & \multicolumn{2}{c}{\textbf{tip error\,{[}mm{]}}} & \multicolumn{2}{c}{\textbf{RMSE\,{[}mm{]}}} \\ 
\text{method} &\text{median} &\text{IQR} &\text{median} &\text{IQR} \\ \hline
\textbf{M1}  &4.68   &4.29   &2.83  &2.20\\
\textbf{M2}  &6.73   &5.46   &3.97   &2.69\\
\textbf{M3}  &6.85   &5.37   &3.98   &2.69\\
\textbf{M4}  &4.36  &4.46    &2.78   &2.56\\
\textbf{Siamese}  &\textbf{3.11}  &\textbf{3.38}  &\textbf{1.98}  &\textbf{1.97}\\
\hline
    \end{tabular}
    \end{center}
\end{table}

\begin{figure}[!b]
    \centering
    \includegraphics[width=0.6\textwidth]{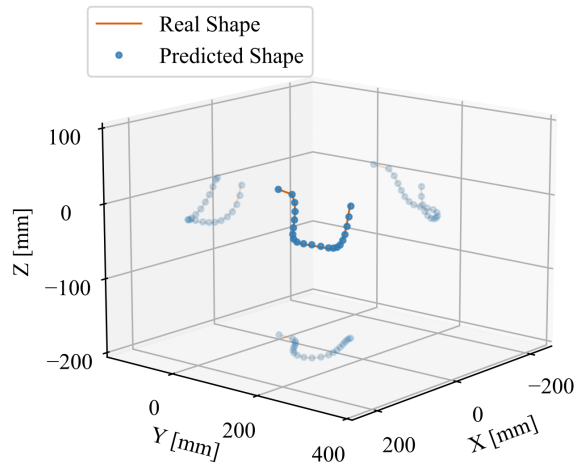}
    \caption{A typical case of predicted shapes using the designed Siamese model. The transparent lines and circles represent the $XY$, $XZ$, and $YZ$ projections in the 3D graphs.}
    \label{fig:3Dshape}
\end{figure}

\begin{figure}[!t]
    \centering
    \includegraphics[width=1\textwidth]{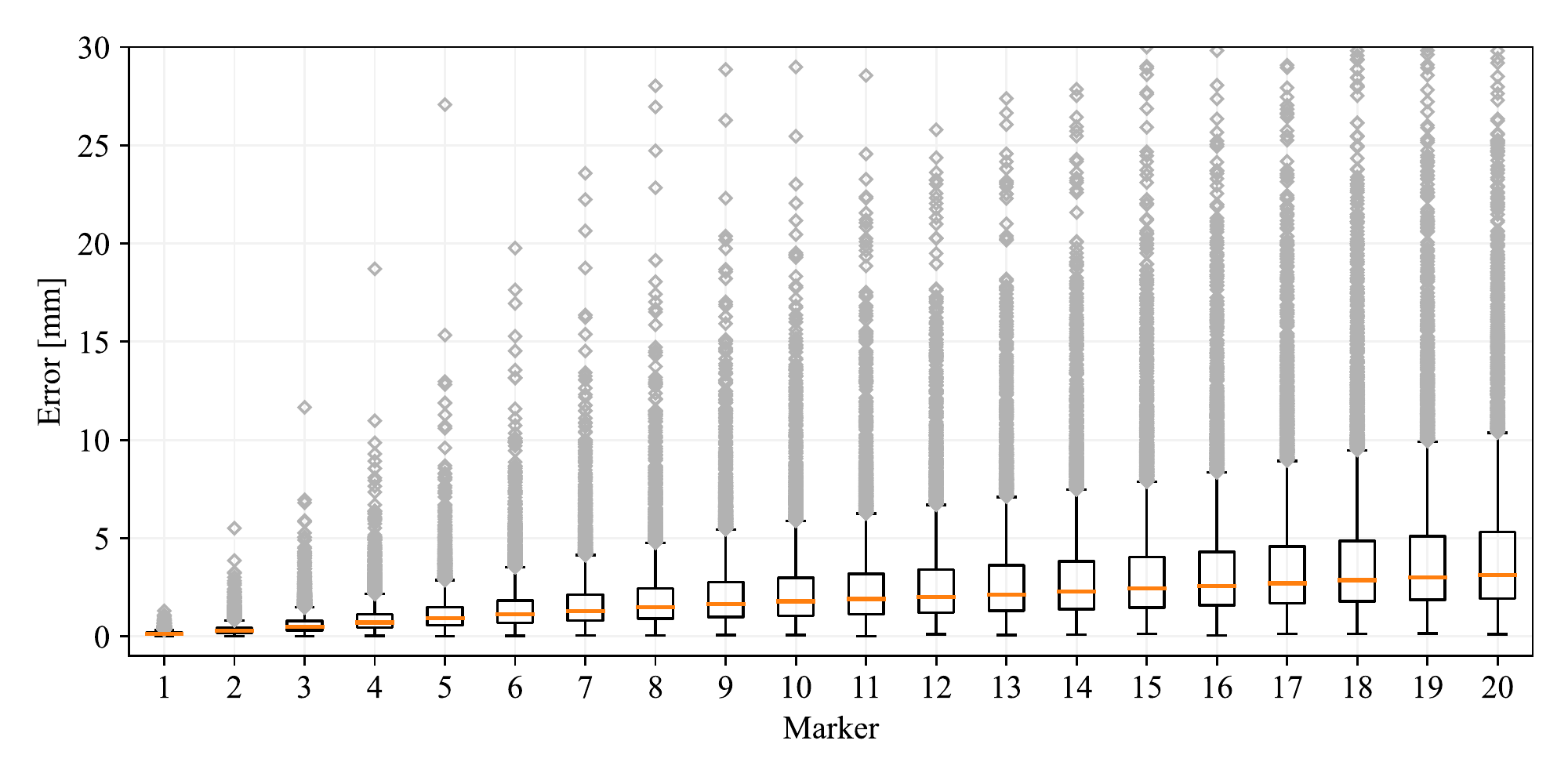}
    \caption{The error statistics are based on the Euclidean distance between the true markers' absolute coordinates and the predicted relative coordinates that are scaled back to the absolute values (the first marker is excluded). On each box, the median is indicated with a central mark, and the $25^{\text{\tiny th}}$ and $75^{\text{\tiny th}}$ percentiles are indicated with the bottom and top edges of the box, respectively. The whiskers show the minimum and maximum values within each group not considered outliers, and the outliers are marked with $\diamond$ symbols. The total number of outliers are $\sim$\,\SI{400} out of $\sim$\,\SI{5300} samples. Some outliers were excluded for better visualization of the median values.}
    \label{fig:Euclidean}
\end{figure}

The designed Siamese network uses the RMSprop as the optimizer with a learning rate of $1^{-4}$, a momentum of $0.9$, and a decay factor of $0.7$. The loss function’s hyperparameters, including $M$, $\alpha$ and $\delta$, are 0.5, 0.7, and 2.2, respectively. 
A typical case of predicted shapes using the designed Siamese network is shown in Figure~\ref{fig:3Dshape}. The error statistics based on the Euclidean distance between the true and the predicted shapes are shown in Figure~\ref{fig:Euclidean}. It can be noticed that the median of the Euclidean distance between the true and the predicted shapes is increasing towards the sensor's end. As explained in \cite{manavi2021using}, this accumulative error might be due to inaccuracies in predicting the sensor's initial orientation. 

\section{Conclusion}
\label{sec:Conclusion}
In this work, we designed a deep learning-based model to extract the shape information of an edge-FBG sensor based on its full spectrum. We used the Hyperband algorithm to optimize the hyperparameters of our neural networks. We performed the hyperparameter tuning in two steps to avoid a large search space. First, the parameters related to the conv1D and the pooling layers were optimized. The best-performing architecture contains seven conv1D layers with a Sigmoid activation function and four max pooling layers. In the second step, optimization and loss calculation hyperparameters were defined in the Hyperband search space to optimize the networks with differently scaled input and output data. We showed that the models perform better when the input data are normalized with the one-dimensional z-scaling and when relative coordinates instead of absolute values are used as the target data. Upgrading the selected architecture to the Siamese design significantly improved the shape prediction accuracy of a \SI{30}{\cm} long sensor, with a median tip error of \SI{3.11}{\mm} and a median RMSE of \SI{1.98}{\mm}.
We achieved an improvement of almost \SI{2.7}{\mm} in the median value of the tip error, and \SI{1.4}{\mm} in the median value of the RMSE using the proposed model compared to the previously designed network architecture proposed in \cite{manavi2021using} which showed a median tip error of \SI{5.8}{\mm} and a median RMSE value of \SI{3.4}{\mm}. Compared to the mode field dislocation method~\cite{roodsari2022secretarxive,waltermann2018multiple}, our proposed model can accurately predict the sensor's shape with an order of magnitude lower tip error.

In future work, we will add temporal shape information to the input data to further improve the prediction accuracy. We also tend to continue investigating different architectural designs, including Siamese networks with triplet loss. 

\section*{Supplementary information}
In the supplementary material, we provided a video of the sensor's predicted shapes using the designed Siamese model.

\section*{Acknowledgement}
We gratefully acknowledge the funding of this work by the Werner Siemens Foundation through the MIRACLE project.

\bibliographystyle{elsarticle-num} 
\bibliography{EdgeFBG_paper}

\appendix
\subsection{Hyperparameter Optimization}
This section presents the search criteria for all three hyperparameter optimizations performed in this work. Table~\ref{tab:Searchingcriteria1} shows the search space settings for the first step of hyperparameter optimization, in which the number of conv1D layers was set to seven. Each conv1D layer was followed by a Sigmoid activation function and batch normalization. For the second step of hyperparameter optimization, the settings of the conv1D and the pooling layers were fixed. Table~\ref{tab:Searchingcriteria2} shows the search space settings for this hyperparameter tuning. The search criteria for tuning the Siamese network's hyperparameters are presented in Table~\ref{tab:Searchingcriteria3}.

\begin{table}[!h]
    \fontsize{10}{12}\selectfont
    \caption{Search criteria for the first step of hyperparameter optimization. The optimized kernel size, channel sizes, and pooling layer settings, resulting from this hyperparameter tuning step, are fixed in the next hyperparameter search step.}
    \label{tab:Searchingcriteria1}
    \begin{tabular}{ll}
    \hline
    \textbf{hyperparameter} & \textbf{search space} \\ \hline
    dropout rate & min: 0, max: 0.3, step: 0.1 \\
    optimizer   & SGDW, Adamw \\
    learning rate   & 0.1, 0.01, 0.001, 0.0001 \\ 
    weight decay   & 0.1, 0.01, 0.001, 0.0001, 0.00001 \\
    momentum    & min: 0, max: 0.9, step: 0.1 \\
    kernel size (similar for all conv1D layers)    & min: 2, max: 10, step: 1 \\ 
    channel size (different for each conv1D layer)  & min: 8, max: 256, step: 8 \\
    choice of max pooling (different after each conv1D layer) & true, false \\
    pooling size (different for each max pooling layer)    & min: 2, max: 3, step: 1 \\
 \hline
    \end{tabular}
\end{table}

\begin{table}[!h]
    \fontsize{10}{12}\selectfont
    \caption{Search criteria for the second step of hyperparameter optimization. In this step, the hyperparameter search is performed three times for each data preprocessing approach.}
    \label{tab:Searchingcriteria2}
    \begin{tabular}{ll}
    \hline
    \textbf{hyperparameter} & \textbf{search space} \\ \hline
    dropout rate & min: 0, max: 0.3, step: 0.1 \\
    optimizer   & SGDW, Adamw, RMSprop, Adadelta, Adamax \\
    learning rate   & 0.1, 0.01, 0.001, 0.0001 \\ 
    weight decay   & 0.1, 0.01, 0.001, 0.0001, 0.00001 \\
    momentum    & min: 0, max: 0.9, step: 0.1 \\
    loss function   &\begin{tabular}[l]{@{}l@{}}mean absolute error, mean squared error,\\mean squared logarithmic error, huber loss,\\ mean absolute percentage error, cosine similarity\end{tabular}
     \\\hline
    \end{tabular}
\end{table}

\begin{table}[!h]
    \fontsize{10}{12}\selectfont
    \caption{Hyperparameter search criteria for Siamese network. $\alpha$, $\delta$, and $M$ are the loss function's hyperparameters, defined in Eq.~\ref{eq:loss}.}
    \label{tab:Searchingcriteria3}
    \begin{tabular}{ll}
    \hline
    \textbf{hyperparameter} & \textbf{search space} \\ \hline
    $\alpha$    & min: 0, max: 1, step: 0.1 \\
    $\delta$    & min: 0.1, max: 5, step: 0.1\\
    $M$    & min: 0.5, max: 1, step: 0.1 \\
    rho    & min: 0.5, max: 0.9, step: 0.1 \\
    momentum    & min: 0.5, max: 0.9, step: 0.1 \\
    \hline
    \end{tabular}
\end{table}

\subsection{Point Cloud Rescaling}\label{PointCloudRescaling}
\begin{figure}[!b]
    \centering
    \includegraphics[width=0.6\textwidth]{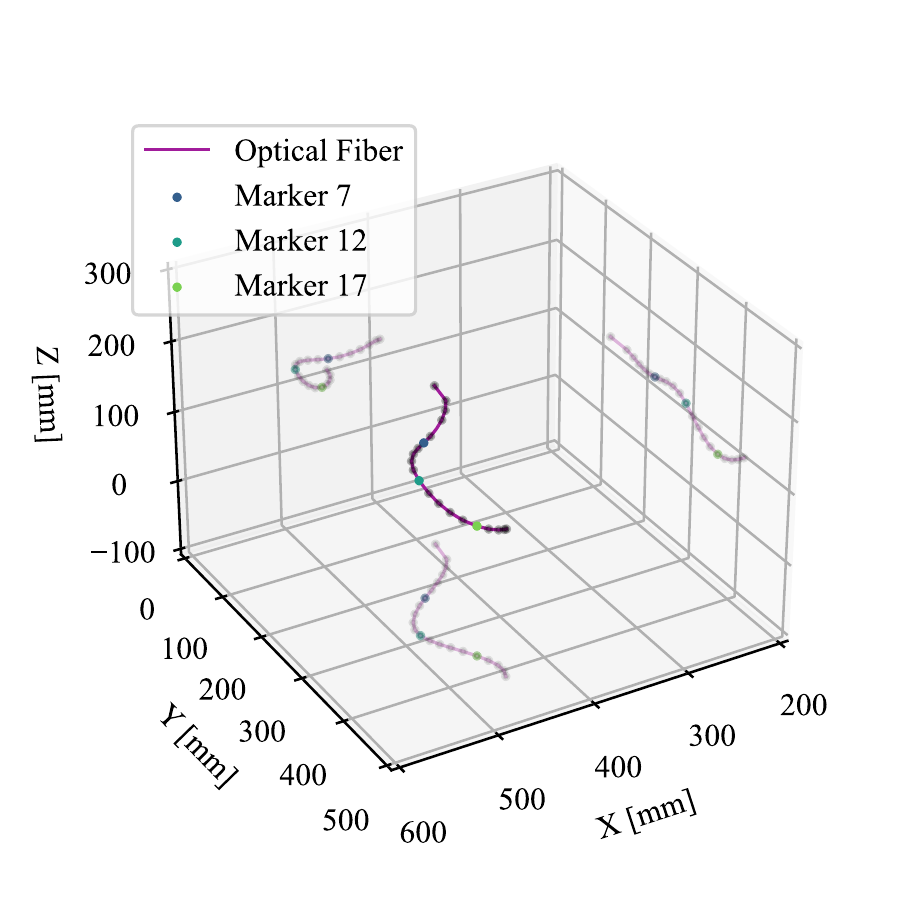}
    \caption{A 3D shape example of the fiber sensor with 21 attached reflective markers is depicted. The three highlighted markers, numbers 7, 12, and 17, are selected for scatter plotting. The remaining  reflective markers are shown with gray circles. }
    \label{fig:3Dexample}
\end{figure}

\begin{figure}[!b]
    \centering
    \includegraphics[width=1\textwidth]{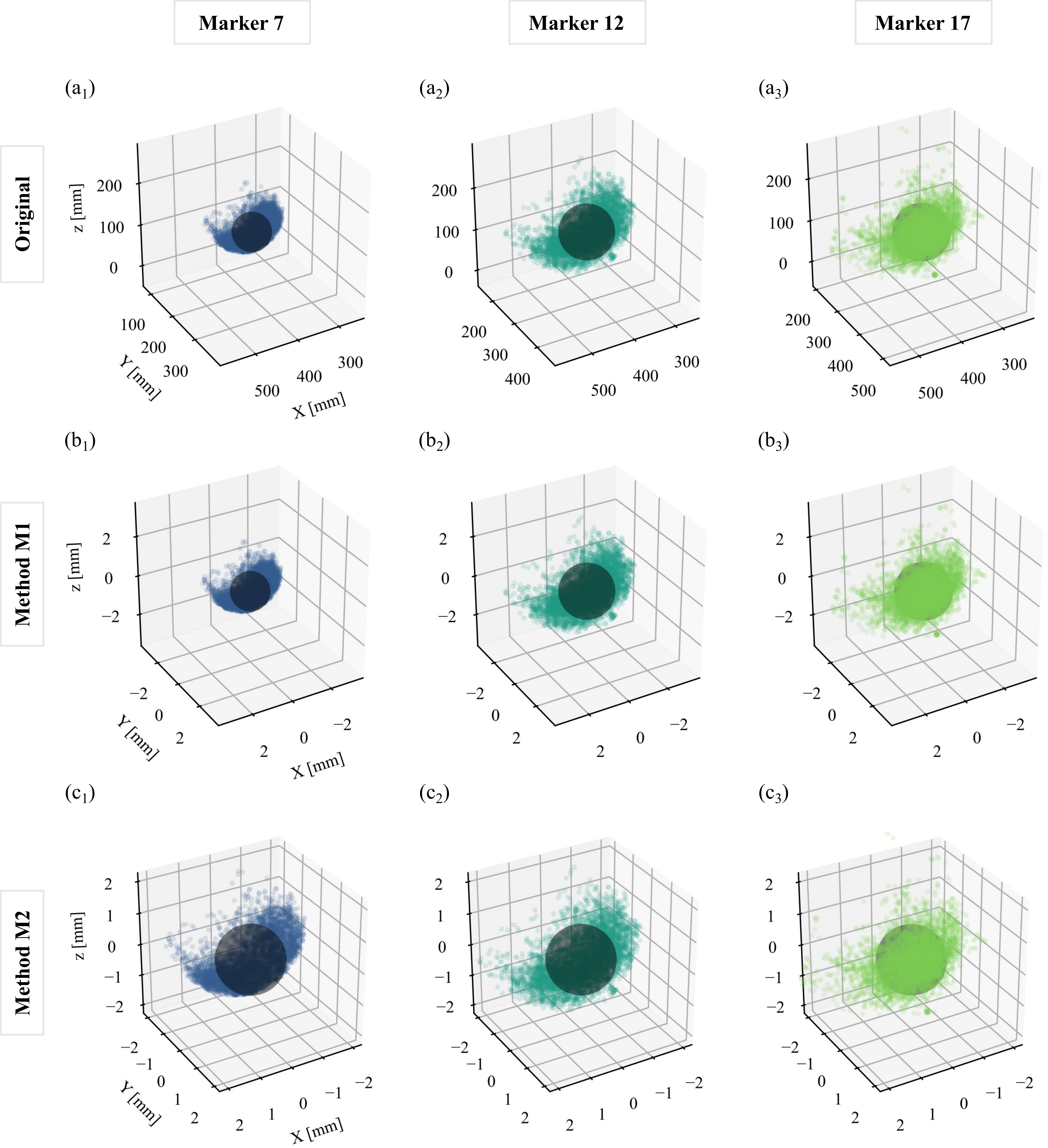}
    \caption{The scatter plot of measured coordinates in $\sim$ 5800 shape examples. The estimated sphere of each point cloud is shown with a black mesh plot. The Original coordinate values of marker numbers 7 ($a_{\text{\tiny 1}}$), 12 ($a_{\text{\tiny 2}}$), and 17 ($a_{\text{\tiny 3}}$). The rescaled point clouds using M1 method for marker numbers 7 ($b_{\text{\tiny 1}}$), 12 ($b_{\text{\tiny 2}}$), and 17 ($b_{\text{\tiny 3}}$). The rescaled point clouds using M2 method for marker numbers 7 ($c_{\text{\tiny 1}}$), 12 ($c_{\text{\tiny 2}}$), and 17 ($c_{\text{\tiny 3}}$).}
    \label{fig:OriginalM1M2}
\end{figure}

\begin{figure}[!h]
    \centering
    \includegraphics[width=1\textwidth]{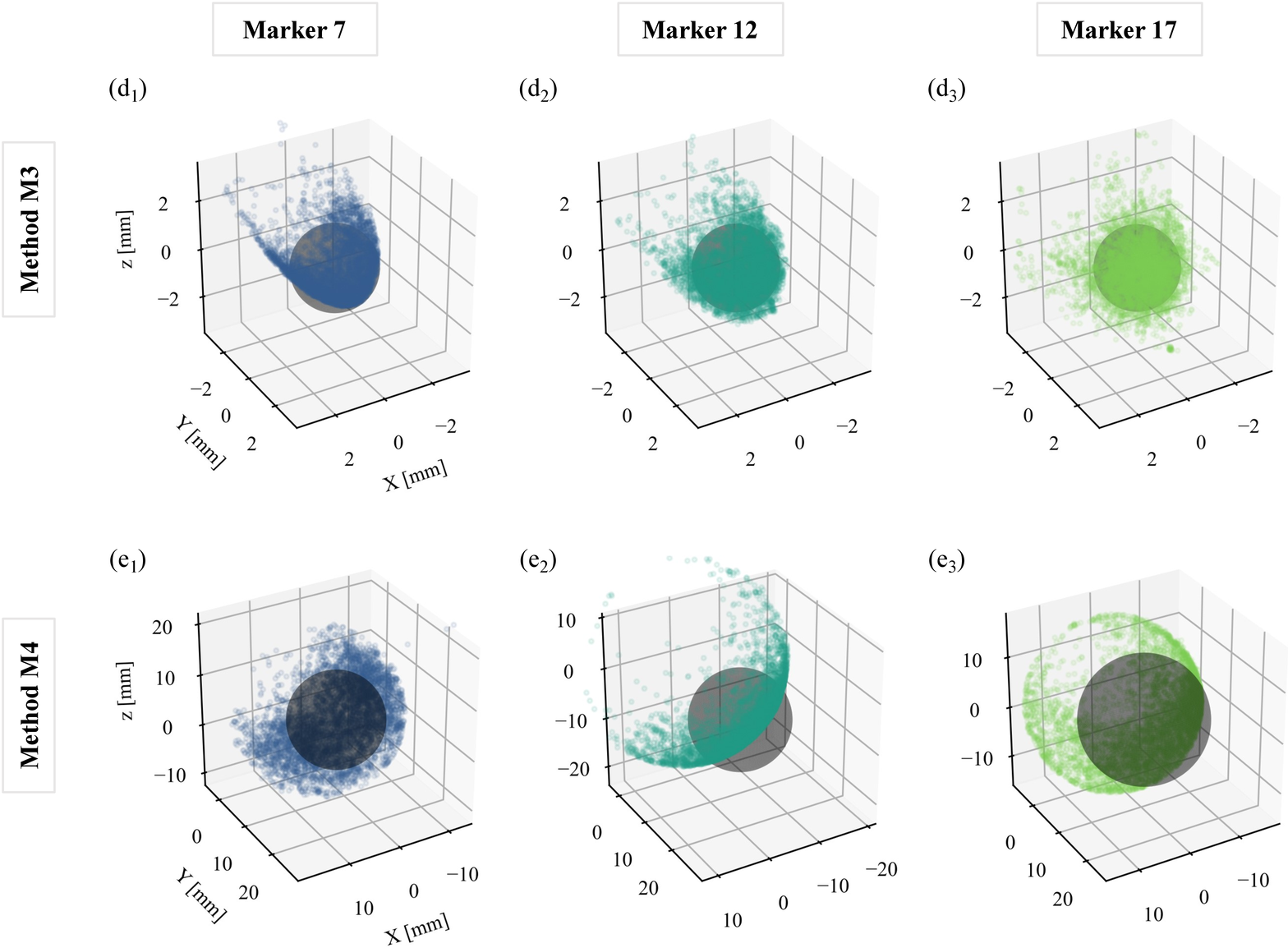}
    \caption{The rescaled point clouds using M3 method for marker numbers 7 ($d_{\text{\tiny 1}}$), 12 ($d_{\text{\tiny 2}}$), and 17 ($d_{\text{\tiny 3}}$).The rescaled point clouds using M4 method for marker numbers 7 ($e_{\text{\tiny 1}}$), 12 ($e_{\text{\tiny 2}}$), and 17 ($e_{\text{\tiny 3}}$).}
    \label{fig:M3M4}
\end{figure}

This section explains the output data transformation approaches in more detail. Figure~\ref{fig:3Dexample} shows a 3D shape example of the fiber sensor with 21 markers. For illustration purposes, we only focus on three highlighted markers (numbers 7, 12, and 17) and study their point cloud modifications as we apply M1-M4 rescaling methods. Figures~\ref{fig:OriginalM1M2} and \ref{fig:M3M4} depict the scatter plots (point clouds) of measured coordinates of the selected markers in $\sim$ 5800 different shape examples. To compare the size of the point clouds after rescaling, an estimated sphere of each point cloud is depicted with a black mesh plot. Each sphere is centered at the mean coordinate of the original/rescaled points and has a radius of $r$. The radius $r$ in each point cloud is the mean value of the calculated radial distance between the points and the cloud's center.

In method M1, the rescaling factor is the same for all spheres. Therefore, as can be seen in Figures~\ref{fig:OriginalM1M2} ($b_{\text{\tiny 1-3}}$), the relative size between the spheres is similar to their original versions (Figures \ref{fig:OriginalM1M2} ($a_{\text{\tiny 1-3}}$)). In method M2, the rescaling factor for each point cloud is the average of calculated $r$ values in that point cloud. Therefore, all point clouds have sphere mesh plots with a radius of one (Figures~\ref{fig:OriginalM1M2} ($c_{\text{\tiny 1-3}}$)).
In the third method (M3), applying three-dimensional standard scaling to the point clouds makes the transformed data uncorrelated (Figures~\ref{fig:M3M4} ($d_{\text{\tiny 1-3}}$)). Using relative coordinates between the markers (M4) greatly changes the point clouds' appearance. As the markers are fixed on the sensor, the maximum relative distance between two neighboring markers is limited. It can be seen in Figures~\ref{fig:M3M4} ($e_{\text{\tiny 1-3}}$) that the coordinate points, especially for the markers closer to the sensor's tip, are constrained in terms of volume and better form a sphere.

\subsection{Significance Test}
\begin{figure}[!h]
    \centering
    \includegraphics[width=0.9\textwidth]{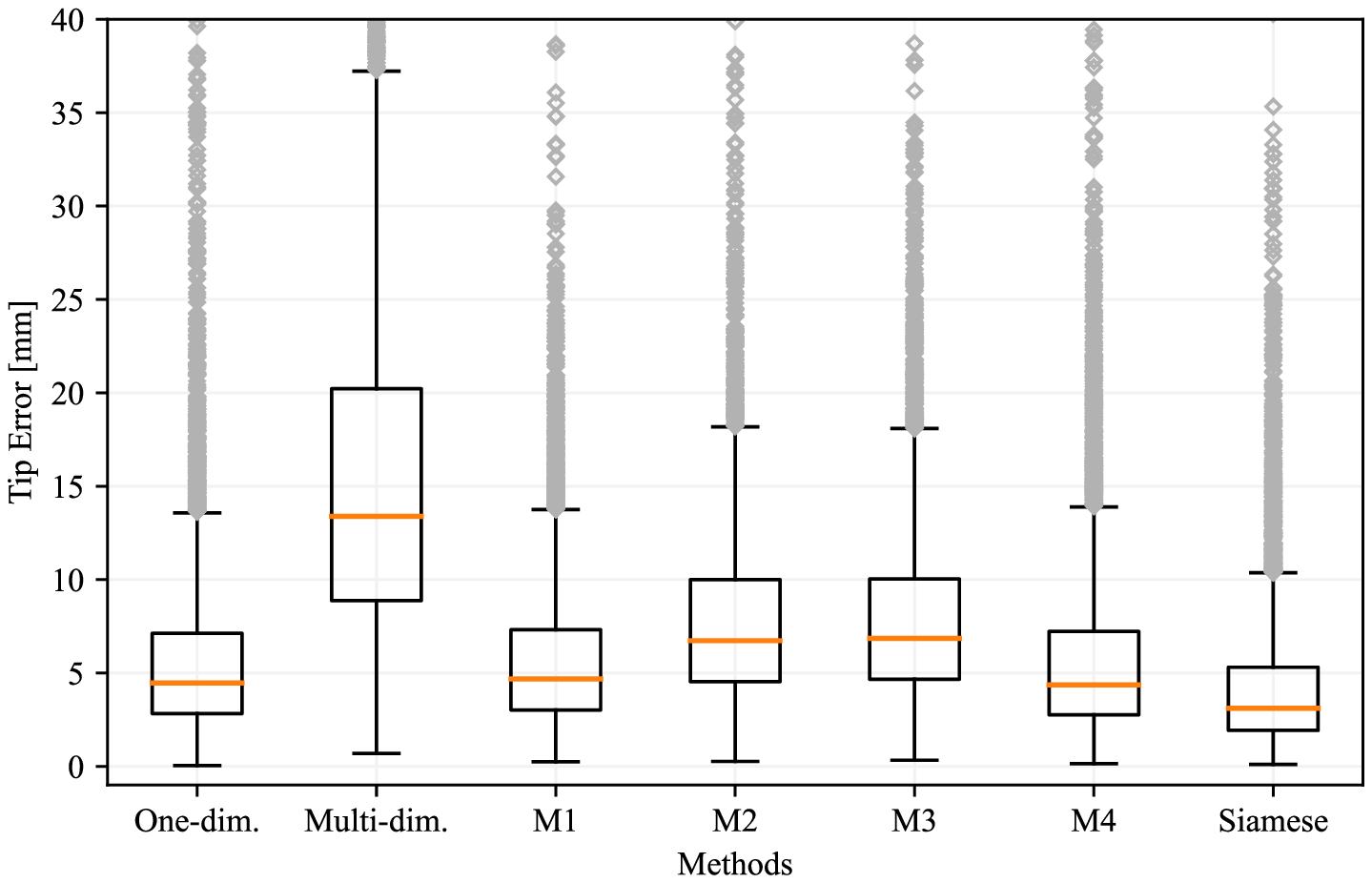}
    \caption{The tip error statistics of the designed networks when different data preprocessing methods and the Siamese architecture were employed. On each box, the median is indicated with a central mark, and the $25^{\text{\tiny th}}$ and $75^{\text{\tiny th}}$ percentiles are indicated with the bottom and top edges of the box, respectively. The whiskers show the minimum and maximum values within each group not considered outliers, and the outliers are marked with $\diamond$ symbols. Some outliers were excluded for better visualization of the median values.}
    \label{fig:signif}
\end{figure}

In this section, the significance test results, comparing the evaluated models, are presented. Figure~\ref{fig:signif} shows the tip error box plots of all seven methods evaluated in this paper. As can be clearly noticed, the Siamese network has the least median tip error. We performed Tukey's HSD pairwise group comparisons on the seven methods. The Siamese method shows p-values close to zero compared to the other seven methods, proving that the shape prediction's improvement is statistically significant. 

\end{document}